# GSAE: an autoencoder with embedded gene-set nodes for genomics functional characterization

Hung-I Harry Chen[1,2], Yu-Chiao Chiu[2], Tinghe Zhang[1], Songyao Zhang[1,4], Yufei Huang[1,§], Yidong Chen[2,3,§]

[1]Department of Electrical and Computer Engineering, The University of Texas at San Antonio, San Antonio, TX 78249, USA

[2]Greehey Children's Cancer Research Institute, The University of Texas Health Science Center at San Antonio, San Antonio, TX 78229, USA

[3]Department of Epidemiology & Biostatistics, The University of Texas Health Science Center at San Antonio, San Antonio, TX 78229, USA

[4]Laboratory of Information Fusion Technology of Ministry of Education, School of Automation, Northwestern Polytechnical University, Xi'an, Shaanxi, 710072, China

[§]Corresponding authors

Email addresses:
  YH: yufei.huang@utsa.edu
  YC: cheny8@uthscsa.edu

# Abstract

**Background**

Bioinformatics tools have been developed to interpret gene expression data at the gene set level, and these gene set based analyses improve the biologists' capability to discover functional relevance of their experiment design. While elucidating gene set individually, inter gene sets association is rarely taken into consideration. Deep learning, an emerging machine learning technique in computational biology, can be used to generate an unbiased combination of gene set, and to determine the biological relevance and analysis consistency of these combining gene sets by leveraging large genomic data sets.

**Results**

In this study, we proposed a gene superset autoencoder (GSAE), a multi-layer autoencoder model with the incorporation of *a priori* defined gene sets that retain the crucial biological features in the latent layer. We introduced the concept of the gene superset, an unbiased combination of gene sets with weights trained by the autoencoder, where each node in the latent layer is a superset. Trained with genomic data from TCGA and evaluated with their accompanying clinical parameters, we showed gene supersets' ability of discriminating tumor subtypes and their prognostic capability. We further demonstrated the biological relevance of the top component gene sets in the significant supersets.

**Conclusions**

Using autoencoder model and gene superset at its latent layer, we demonstrated that gene supersets retain sufficient biological information with respect to tumor subtypes and clinical prognostic significance. Superset also provides high reproducibility on survival analysis and accurate prediction for cancer subtypes.

# Background

Nowadays gene set based analysis has been an essential step for interpreting gene expression data, for which a variety of bioinformatics tools have been developed to extract biological insights from different aspects. Among all methods, functional enrichment is the most common gene set based analysis to determine classes of genes that are associated with disease phenotypes, such as Gene Set Enrichment Analysis (GSEA) [1]. Function over-representation is another category for enrichment analysis, represented by The Database for Annotation, Visualization and Integrated Discovery (DAVID) [2, 3], among many others [4-6]. Researchers also employ gene set as a classifier; for example, the 50-gene PAM50 model was used to classify the subtypes of breast cancer [7]. Furthermore, many studies have conducted survival analysis at gene set level to predict clinical outcomes [8, 9]. Overall, gene set analysis improves the biologists' capability to interpret functional impact to their experiment design. However, some studies have also disclosed the inconsistency of gene set results. Lau *et al.* showed that there are only minimal overlaps between the putative prognostic gene sets for non-small-cell lung cancer found in nine various studies [10]. Hence, inter gene sets association should be taken into consideration, as suggested by various studies, to limit inconsistency. While combined gene sets may provide consistency, its biological relevance are rarely discussed.

Deep learning methods have emerged recently in computational biology due to the increase of molecular and cellular profiling data. Convolutional neural network (CNN) methods were implemented for prediction of DNA-protein binding [11] or detection of phenotype-associated cell subsets [12]. Autoencoder, which is an unsupervised learning algorithm, was used for modeling gene expression through dimensionality reduction in many studies [13-15]. Lin *et al.* proposed a supervised neural network model for single-cell RNA-seq data that incorporate protein–protein interaction (PPI) and protein–DNA interaction (PDI) information. However, the prior biological knowledge was only utilized to improve the performance of dimensionality reduction and cell type-specific identification, the influence of combining PPI nodes was not examined.

In this study, we proposed Gene Superset AutoEncoder (GSAE), a multi-layer autoencoder model that incorporates *a priori* defined gene sets to preserve the crucial biological features from combining gene sets in the latent layer. We introduced the concept of the gene superset, an unbiased combination of gene sets, with weights trained by the autoencoder, where each node in the latent layer is termed a superset. The goal of this study is to determine the functional or clinical relevance of the learned gene supersets from our model, where the model evaluates gene expression data at the level of superset. To achieve our goal, we used large-scale RNA-seq data sets from The Cancer Genome Atlas (TCGA) to test GSAE and investigate the top ranked gene sets in the statistically significant supersets. We demonstrated that gene supersets preserve sufficient biological information with respect to tumor subtypes and clinical prognostic significance. Our study also compared different neural network classifiers and the superset classifier showed high accuracy in cancer subtype prediction. We concluded that superset produces more reproducible results than single gene sets, provides robustness in cancer subtype classification, and has the capability to learn potential gene sets association.

## Materials and Methods

### Data sets in this study

For Pan-cancer (PanCan) analysis, we collected TCGA RNA-seq data that was organized by TumorMap [16], which contains 9,806 samples in 33 cancer types. In addition to entire TCGA data, we also selected breast invasive carcinoma (BRCA) data with 1,099 samples for characterizing network nodes. For survival analysis, lung adenocarcinoma (LUAD) with 515 samples were chosen. Furthermore, we used four data sets with sufficient survival information, LUAD, BRCA, lower grade glioma (LGG, 523 samples), and skin cutaneous melanoma (SKCM, 469 samples) to compare the reproducibility of supersets and gene sets. The expression profiles of all tumor RNA-seq in this study are in the Transcripts Per Million (TPM) unit and then log-transformed (logTPM = $log2$(TPM + 1)), which are re-analyzed uniformly for all samples [16].

**Gene superset autoencoder**

The architecture of GSAE is shown in Fig. 1. The input of the model is the gene expression profiles in log2 TPM values. The output $x$ of the $j$th node in the $i$th layer can be formulated as

$$x_{ij} = g\left(b_{(i-1)} + \sum_j w_{(i-1)j} \, x_{(i-1)j}\right) \quad (1)$$

where the bias $b$ and the weight $w$ are the two parameters that are learned in training, $g()$ is the activation function, where we used the linear activation in the output layer and rectified linear unit (ReLU, defined in Eq. 2) in other layers to provide non-linearity while keeping a scoring feature in the model.

$$ReLU = \begin{cases} x, if\ x > 0 \\ 0, otherwise \end{cases} \tag{2}$$

Besides the input layer in our proposed autoencoder, the first two layers are the encoding network that extracts the important features from gene expression. ReLU activation mimics pathway activation/deactivation function, and comparing with with linear activation in all layers, GSAE with ReLU activation in the hidden layers has much better performance in cancer subtype prediction (data not shown). The decoder part comprises the last two layers; it is a complementary function of the encoder, and it aims to reconstruct the input from the converge layer. If the model is designed as a neural network-based classifier for cancer subtype prediction, the decoder network is replaced by a softmax function that is used as the output layer. At last, we choose the loss function to be either a mean square error function for the reconstruction loss, or a categorical cross-entropy function for multi-class classification error.

**Incorporate gene sets into the encoder layer**

We designed the first layer in the encoder as the gene set layer, which incorporates the information of a collection of gene sets. Specifically, each node in this layer represents a gene set, where only genes in the input layer that belong to a gene set have connection to the node [13], and the weight for each connection is determined by the backpropagation in training steps. This is different from the fully connected layer commonly used in autoencoder. We adopted the chemical and genetic perturbations (CGP) collection downloaded from the Molecular Signatures Database (MSigDB) [1, 17] and replaced some highly dependent gene sets with a representative gene set among them.

From the output of the gene set layer, we can retrieve the gene set score of each input sample. Following with a dimension reduced dense layer, the superset layer (latent layer), can be used to investigate the combination of gene sets while keeping the crucial features needed to reconstruct the input data by the decoder. The superset layer is the key layer of our model, which we obtain a group of gene sets that correlate with divergence of cancer subtypes. Each node in this layer is considered as a superset, which is a combination of different gene set terms. In this study, we set the superset layer size to 200. With the information of supersets, we can analyze characteristics of the data set, such as development of subpopulations or clinical relevance of a disease.

**Resolve dependencies among gene sets**

The CGP collection in MSigDB includes the gene sets that represent expression signatures of genetic and chemical perturbations published in literature. However, some gene sets are highly similar, and we need to mitigate the dependency. We used a similar method as in our previous study [18] to cluster gene sets with significant similarity. First, we omitted the gene sets that have less than 15 or more than 500 genes, which is also the default setting in original GSEA implementation [1]. We subsequently used kappa statistics to measure the similarity between all gene sets. We clustered gene sets with *P*-value $< 10^{-7}$, and assigned the largest gene set as the representative of the cluster. At last, there are 2,334 CGP gene sets including 18,107 genes selected to form the gene set layer.

**Establish and train the gene superset autoencoder**

We implemented the model using Keras 1.2.2 (https://github.com/fchollet/keras) and used the custom layer method in Keras to accomplish the sparsity of gene set layer in order to keep the zero weights while optimizing the parameters. Since ReLU is used as the activation function, we selected He uniform initialization as the initializers for all layers [19]. To train the autoencoder, we used the stochastic gradient descent (SGD) optimizer that was revised in Lin *et al.* study [13], which was designed to deal with the optimization problem for sparse layers. The SGD parameters were set as following, *learning rate* = 0.05, *decay* = $10^{-6}$, *momentum* = 0.9, and *Nesterov* = 1.

While training the model for a data set, we extracted 5% of data to be the validation set to avoid overfitting. With the *callbacks.EarlyStopping*() function in Keras, the model stops training when the loss of validation split doesn't improve in three consecutive epochs. At last, we imported the data set into the trained model and exported the outputs and weights of the encoder layers for further analyses in R.

**The use of additional machine learning tools**

In this study, we have applied t-Distributed Stochastic Neighbor Embedding (t-SNE, https://cran.r-project.org/package=Rtsne) [20, 21], which has been widely used for dimensionality reduction. We performed t-SNE on superset results and embedded the high-dimensional data into a two-dimensional space, where potential subpopulations of the data were revealed. Another machine learning method, Hierarchical Density-Based Spatial Clustering of Applications with Noise (HDBSCAN, https://cran.r-project.org/package=dbscan) [22, 23], was used in the tumor subtype analysis. Comparing with many other clustering algorithms, HDBSCAN has good performance and stability in exploratory data analysis. We performed HDBSCAN on the t-SNE results to determine the possible clusters among the data. Ambiguous samples were classified as noise and omitted from further analysis.

**Differential superset analysis between tumor subtypes**

After performing t-SNE on the superset layer outputs, we subsequently determined the subtypes of a data set by using HDBSCAN. To find the supersets with a subtype pattern, we compared superset values between one tumor subtype (group 1) and the other subtypes (group 2) by one-tailed Mann-Whitney-Wilcoxon *U* test (MWW) with a location shift of "*mu*" (*mu* was assigned to change the stringency of the test). Significant supersets (MWW *P*-value < 0.01) that have larger values in group 1 were named as up-supersets, whereas down-supersets were the significant supersets with larger than in group 2. We further investigated gene sets in the significant supersets. To quantify the contribution of $i^{th}$ gene set in $j^{th}$ superset, *gsScore* was calculated as following,

$$gsScore_{ij} = \left(\mu_1^{(i)} - \mu_2^{(i)}\right) \times w_{ij} \qquad (3)$$

where $\mu_1$ and $\mu_2$ are the average of the $i^{th}$ gene set values in the two groups, and $w_{ij}$ is the weight in the model corresponding to the connection from the $i^{th}$ gene set to the $j^{th}$ superset. In up-supersets, gene sets with *gsScore* greater than a positive cutoff (in the right tail) were selected. In the opposite, gene sets in the down-supersets with *gsScore* less than a negative cutoff (in the left tail) were selected. Those gene sets are the potential high impact gene sets of the subtype (group 1).

**Kaplan-Meier Survival analysis on superset layer**

To discover whether GSAE retains survival related features; for each superset and gene set, we used a median split (median of the superset or gene set value) to create two groups and performed log-rank test. For each prognostic significant superset, we ranked gene sets according to the *gsScore* (Eq. 3) and further investigated the survival relevance of top gene sets.

## Results

**Cancer type information Preserved in low dimension outcome**

To test the capability of GSAE to retain crucial features in the superset layer, we used TCGA PanCan RNA-seq logTPM data, 15,975 genes selected with $\mu > 1$ and $\sigma > 0.5$ across 9,806 samples in 33 cancer types, as GSAE inputs and exported the superset layer results. We performed t-SNE on TCGA logTPM data and the superset layer outputs (200 nodes), and the results are shown in Fig. 2, in which the color of each node was labelled according to the cancer type information. The groupings of cancer types in the two t-SNE plots are nearly identical, where most cancer types form an individual cluster. The mingling of few cancer types are also similar in both figures. We concluded that the model is able to retain cancer type features of a data while reducing dimensionality.

**Indication of gene sets associated with breast cancer subtypes**

In Fig.2, we learned that the samples labeled in red are separate into two clusters, and we further verified that they belonged to BRCA. We used GSAE to analyze the BRCA data separately to discover gene sets that are support this subtype differentiation. There were 15,183 genes in 1,099 samples that meet the criterion of $\mu > 1$ and $\sigma > 0.5$, where they were used as the model input. After training of the model, we exported the superset results and performed t-SNE, which is shown in Fig. 3A. We applied HDBSCAN, which clustered the samples into two groups, where group 1(G1) is labeled in red and group 2 (G2) in green. The noisy samples defined by the algorithm were omitted. Four up-supersets and three down-supersets were determined (*P*-value < 0.01) using one-tailed Mann Whitney *U* test with location shift *mu* = 9, where only supersets with a huge difference between the two groups could pass the test. In each significant superset, those with *gsScore* > 2 *sd* (standard deviation of all *gsScores* in the superset) are the high impact gene sets of the superset. All high impacts gene sets of 7 significant supersets are listed in table S1, and the *PScore* ($-\log_{10}(P\text{-value})$) of Mann Whitney *U* test (location shift set as 0.5) of each gene set was also included.

Top 15 gene sets in the most significant up-superset and down-superset are listed in Tables 1 and 2. The two superset density plots of gene set values (output of gene set nodes) in Fig. 3B and 3C show the vast difference between the two groups in those significant supersets. We also identified many high impact gene sets associated with breast cancer subtypes. For example, "FARMER_BREAST_CANCER_BASAL_VS_LULMINAL" clearly indicates the two groups are possible Basal and Luminal subtypes [24]. The study of "STEIN_ESRRA_TARGETS_RESPONSIVE_TO_ESTROGEN_UP" gene set also suggested that *ESRRα* might be a therapeutic target for triple negative breast cancer [25]. Group 1 has a higher value in "DOANE_BREAST_CANCER_ESR1_DN", which matches the gene set condition where these genes were down-regulated in *ESR1* positive samples [26]. Genes that are involved in "PEDERSEN_METASTASIS_BY_ERBB2_ISOFORM_7" differentiates the HER2 positive and negative BRCA subtype [27]. A study has shown that *c-Myb* differed significantly across the subtypes, where Basal-like has the lowest expression [28], which fits the result of "LEI_MYB_TARGETS". There is no direct connection of Estradiol with breast cancer subtype, but it is an estrogen and its target gene set "FRASOR_RESPONSE_TO_ESTRADIOL_DN" might be a potential subtype marker.

After reviewing these gene sets, because the Basal subtype accounts for 15% of the breast cancer population, we hypothesized that G1, the small subpopulation in red in Fig. 3A, is the Basal subtype of breast cancer. We checked the TCGA clinical information and PAM50 classification results and verified that 156 of 175samples (with complete estrogen status or PAM50 subtype information) in G1 are either Basal-like or triple negative samples. This result demonstrates that our proposed superset autoencoder is able to reveal the subpopulation features and biological relevance.

We compared with GSEA results between G1 and G2, and 53 out of 124 (42.7%) high impact gene sets are also GSEA enriched gene sets (nom $P$-value < 0.05), which indicates the reliability of our results. To examine whether a superset contains some exclusive gene sets, we compared the top 3 up-supersets (Table S1A-C) and the Venn diagram is shown in Fig. 3D. Many of the overlapped gene sets are associated with the Basal subtype (12 common gene sets in Table S1, bold font). Up-superset 1 has additional estrogen related gene sets (Table S1A, colored in blue); up-superset 2 holds some gene sets that are relevant to *ERBB2* (Table S1B, colored in blue).

**Prediction of breast cancer PAM50 subtypes with superset classifier**

To test if our model can be used as a classifier to predict cancer subtypes, we reconstructed our model to the architecture shown in Fig. S1A, where the decoder network is replaced by a softmax function output (input – encoder – prediction output). With the clinical annotation organized by the UCSC Cancer Genomics Browser [29, 30] (captured in 2015, https://genome-cancer.ucsc.edu), we selected 821 BRCA samples with 15,183 genes in Basal, LumA, LumB, and Her2 PAM50 subtypes as input data to test the performance of the superset classifier (normal-like subtype was removed due to small sample size). Using 10-fold cross validation to test the superset classifier, we achieved a good performance of 88.79% prediction accuracy.

With the same input, we also compared with four different neural network models, 1) gene set classifier, in which the superset layer is removed (Fig. S1B), 2) 2-layer fully connected encoder network with the same size of the superset classifier (Fig. S1C), 3) 2-layer fully connected encoder network, where the size of each layer was optimized by Hyperas [31] (Fig. S1C), and 4) 4-layer fully connected encoder network, where the size of each layer was optimized by Hyperas (Fig. S1D). The size and 10-fold cross validation accuracy of each classifier are listed in Table 3. We have tuned the SGD parameter setting of each model in order to get the best performance.

The prediction accuracy of gene set classifier (87.69%) is close to that of the superset classifier, which implies the gene set layer contains sufficient information for classification. On the other hand, all three classifiers with fully connected encoder has low prediction accuracy (<50%), mainly due to the large number of weights need to be trained to attain (or fail to attain) an optimal model. To alleviate the training burden, we reduced the input number by performing principal component analysis (PCA) on BRCA data first and selected top 500 principal components (PCs) to test the models with fully connected encoder (layer size was also optimized by Hyperas, Table 3). The prediction results (87.57%) are equivalent to the superset classifier, indicating that the gene set layer and top PCs both preserve important subtype features. While both PC classifier and gene set classifier achieved same accuracy, we can design our network to emphasize certain features (e.g. PAM50 subtype classification), based on the fact that we understand the biological functions of a priori defined gene set, and the flexibility of choosing different functional sets (signaling pathways, immunological signatures, etc).

**Prognostic significance for lung adenocarcinoma**

TCGA LUAD data set was employed to test if the model is capable of retaining survival related features in the superset layer. With the same gene selection criterion, 15,188 genes in 515 samples were used as the model input. We also organized the TCGA LUAD survival information to a 5-year survival record, where the maximum survival time was set as 1,825 days, and a death event that occurred after five years was counted as a censored event. After performing log-rank test on the superset results, we determined 6 supersets with log-rank $P$-value < 0.001, which were considered as prognostic significant nodes. We ranked the gene sets in those six supersets by the *gsScore*, and the top 20 gene sets in each superset are listed in Table S2. The top ranked gene sets that also showed significance in gene set log-rank test were selected to probe the biological relevance of lung adenocarcinoma.

We picked the first and fourth ranked supersets as two examples, and the top 15 gene sets in the two supersets are listed in Table 4 and 5. We chose the 4$^{th}$ ranked superset due to the least overlap of significant gene sets with the 1$^{st}$ ranked superset. We selected three gene sets tested significant by the log-rank test from the two supersets and plotted the Kaplan-Meier survival curves in Fig. 4. In rank 1 superset, several significant gene sets are related to the survival of LUAD. A study has shown that decreased mRNA expression of *TCF21*, a tumor suppressor, is a core predictor for poor prognosis in patients with lung cancers in two studies[32-34], agree with what we found the prognosis association from TCGA LUAD with gene set "CUI_TCF21_TARGETS_2_UP" ($P = 1.30 \times 10^{-4}$).

"KIM_WT1_TARGETS_DN" ($P = 0.0064$) is related to the oncogene *WT1* in lung cancer, and the high expression of *WT1* links to an unfavorable impact on the prognosis [35]. We also found some gene sets that no previous study showed direct connection with the prognosis of LUAD. Previous studies have revealed that ETS-related transcription factors are associated with non-small-cell lung cancers (NSCLC) [36, 37]. *ELK3* is also an ETS transcription factor, and the related gene set "GROSS_HYPOXIA_VIA_ELK3_UP" ($P = 5.21 \times 10^{-4}$) might be relevant to LUAD survival. Two chemical compounds related gene sets were discovered in superset 1, "MARTINEZ_RESPONSE_ TO_TRABECTEDIN_DN" ($P = 0.0015$) and "CONCANNON_APOPTOSIS_BY_EPOXOMICIN_DN" ($P = 0.0264$). While both gene lists were derived from studies of other cancer types (e.g., HCT116 colon cancer cell-line), other studies has demonstrated the effectiveness of both Epoxomicin and Trabectedin in lung cancer treatment. Carfilzomib, which is a designed drug based on Epoxomicin, demonstrated anti-proliferative activity and resulted in prolonged survival in mice with SHP-77 small cell lung cancer xenografts [38]. There was only one study testing treatment with trabectedin on NSCLC patients, but no recommendation was given to use trabectedin as single agent treatment [39]. Thus, these two gene sets could be further examined to look for the biological relevance to LUAD.

In 4[th] ranked superset, "IWANAGA_CARCINOGENESIS_BY_KRAS_PTEN_UP" ($P = 0.0249$) is a widely studied gene set to show the association with LUAD prognosis. The study that concluded this gene set observed the aberration in NSCLC with oncogenic form of *KRAS* and inactivated *PTEN*, in which condition resulted shorter survival [40]. The gene set "ZHANG_BREAST_CANCER_PROGENITORS_UP" ($P = 0.0248$) shows the effect of progenitor cells in survival. Ooi *et al.* suggested that the presence of a putative tumor-initiating progenitor cell population in NSCLC is a biomarker with a worse prognosis [41]. *MAPK8* related gene set "YOSHIMURA_MAPK8_TARGETS_DN" ($P = 0.0195$) is also a potential prognostic associated gene set, while only one study implied indirect impact of poor prognosis due to MAPK8 repression [42].

From the two selected supersets, we already found some gene sets highly associated with LUAD survival, there are some novel prognostic gene set candidates that need to be further studied. In conclusion, superset results encompass survival-associated features and sort out the priority of potential prognostic gene sets.

**Improved survival reproducibility from supersets**

To compare the reproducibility of survival results between the superset and gene set layers, we selected four TCGA data sets (BRCA, LUAD, SKCM, and LGG) to examine the reproducibility of GSAE. For each data set, we omitted genes that did not meet the criterion of $\mu > 1$ and $\sigma > 0.5$. We next randomly split 60% of the data as the training set and the remaining 40% as the test set. After the autoencoder was trained on the training set, we obtained the superset outputs for the training and test sets. Median split and log-rank test were performed on training and test superset results to determine survival-related supersets and gene sets.

We assumed that the prognostic significant gene sets and supersets should be similar between training and test data. To evaluate the performance of gene set and superset results, we compared the significant gene sets and supersets obtained from training data and those from test data by Jaccard index. Furthermore, we used two population proportions z-test to examine whether supersets have greater overlap proportion in the training data, and the results are shown in Table 6.

In the largest data set BRCA, we found out that superset has much higher Jaccard index (34%) than gene set (12%), and the two overlap proportions differ significantly ($P = 2\times10^{-4}$). In two other smaller data sets, LUAD and SKCM, superset (Jaccard Index ~18%) still outperforms gene set (11% and 7% for LUAD and SKCM, respectively; z-test P-value < 0.05). In LGG, because of the large number of prognostic significant nodes for superset and gene sets, both Jaccard coefficients are high (~48% for both superset and gene set) and the performance of gene set and superset is identical. To avoid the potential of sampling bias, we repeated the whole process in BRCA and LUAD several times, and we obtained similar stability measure (z-test P-value, data not shown). Overall, we concluded that superset has better reproducibility performance over gene set.

## Discussion

Same as other machine learning algorithms, the selective process of GSAE is an issue. Despite getting identical losses, different nodes (or gene sets) in different training may selective activated or de-activated (output value ~0) with the same training data. Take our study for example, we might obtain the same outcome (e.g. tumor subtype classification) of a dataset in the superset layer, but it is difficult to match superset between runs, and the top ranked gene set components in significant supersets might also be different, although highly relevant gene sets appear more frequent. This observation can be used to assess the significance of a given gene set or superset to a specific aim (e.g. survival association).

We also tried to understand the major cause of selective process in our model, and two possible factors were concluded – the dependency among gene sets in the CGP collection and the initialization of the model weights. Even though we tried to mitigate the dependency effect, many gene sets still share a subset of genes. In the model, we observed that some gene sets with zero values (deactivated) are highly overlapped with top ranked gene sets (activated). We assume that the information (member genes) of a zero-value gene set can be replaced by a combination of other gene sets. In addition, all weights in the GSAE model are randomly initialized. Due to the randomly initialized weights and dependency among gene sets, the model can reconstruct the input data through different gene sets, which results in the selective process of activated or deactivated gene sets. A better choice for independent or less overlapping gene sets could be Gene Ontology (GO) slims, a cut-down version of the whole GO. We might also alleviate this selective issue by assigning saved initial weights from a previous run or pre-trained weights of other data.

Another limitation of our model is the requirement of large sample size, which is a constraint for usual bulk RNA-seq experiments. However, the characteristic of single-cell RNA-seq (scRNA-seq) experiments, lower read depth with much larger sample size (near half-million scRNA-seq in some studies [43, 44]), perhaps is more suitable to be analyzed by deep learning methods. There are many advantages to examine scRNA-seq data using GSAE. First, scRNA-seq analysis with our model will not be restricted by statistical assumptions, where we can avoid dealing with the diverse statistical characteristics of single-cell data [45]. Second, we can directly determine the exclusive gene sets or GO functions of each identified subpopulation, without the need to find the representative genes of a subpopulation. With the support of other additional tools, analyzing scRNA-seq data with our model will be more thorough. For example, using only highly variable genes detected by scVEGs [46] will increase the diversity of subpopulations while lowering the variable dimension of the model. The HDBSCAN algorithm can cluster the multiple subpopulations of single-cell data precisely after t-SNE is applied. Overall, there is a huge potential using deep learning methods for scRNA-seq data analysis.

The concept of gene supersets not only provides better reproducibility, it also gives us a chance to understand the inter-dependency of gene sets. In this study we investigated the associations between significant supersets and gene sets. However, relations between those top ranked gene sets in the same superset has yet to be discussed. One possible solution is to find the corresponding input genes that have large contribution to a significant superset (by interpreting the weights in the first layer as the gene weights in each gene set), where we can further form a set of genes based on the superset. All these alternative approaches will guide our future study to bolster the biological functions of supersets.

## Conclusions

In this paper, we proposed a multi-layer autoencoder model with the incorporation of annotated gene set information. The model is capable of preserving crucial biological features of gene expression data in the dimension reduced superset layer. From the superset results, we have found out information such as tumor subtype differentiation and clinical prognostic significance. With the concept of superset, an unbiased combination of gene sets, we can improve the reproducibility of survival analysis, provide robust prediction of cancer subtypes, and indicate potential gene sets association of a disease. GSAE has the versatility to incorporate different gene set collection, discover different biological relevance, and analyze different kinds of gene expression data.

## List of abbreviations

GSAE: Gene Superset Autoencoder

GSEA: Gene Set Enrichment Analysis

DAVID: The Database for Annotation, Visualization and Integrated Discovery

CNN: Convolutional neural network

PPI: Protein–Protein Interactions

PDI: Protein–DNA Interactions

TCGA: The Cancer Genome Atlas

PanCan: Pan-Cancer

BRCA: breast invasive carcinoma

LUAD: lung adenocarcinoma

LGG: lower grade glioma

SKCM: skin cutaneous melanoma

TPM: Transcripts Per Million

ReLU: rectified linear unit

CGP: chemical and genetic perturbations

MSigDB: Molecular Signatures Database

SGD: Stochastic Gradient Descent

t-SNE: t-Distributed Stochastic Neighbor Embedding

HDBSCAN: Hierarchical Density-Based Spatial Clustering of Applications with Noise

MWW: Mann-Whitney-Wilcoxon *U* test

NSCLC: non-small-cell lung cancers

GO: Gene Onotology

PCA: Pricipal Component Analysis

scRNA-seq: single-cell RNA-seq

## Authors' contributions

All authors contribute to the manuscript. HHC, YCC, TZ, SZ, YH and YC conceived and designed the study. HHC carried out the Python and R scripts. HCC and YCC investigated the biological relevance of the results. All authors read and approved the final manuscript.

## Ethics approval and consent to participate

Not applicable

## Consent for publication

Not applicable

## Availability of data and materials

Data will be shared upon request.

## Acknowledgement

Funding for this research was provided partially by the NCI Cancer Center Shared Resources (NIH-NCI P30CA54174 to YC), NIH (CTSA 1UL1RR025767-01 to YC, R01GM113245 to YH), CPRIT (RP160732 to YC), and San Antonio Life Sciences Institute Innovation Challenge Award 2016 to YC and YH.

## Competing interests

Authors declare no competing interest in preparing the paper and developing the software associated to this paper.

## Declaration

The publication costs for this article were funded by the aforementioned CPRIT grants to YC.

# Reference

1. Subramanian A, Tamayo P, Mootha VK, Mukherjee S, Ebert BL, Gillette MA, Paulovich A, Pomeroy SL, Golub TR, Lander ES *et al*: **Gene set enrichment analysis: a knowledge-based approach for interpreting genome-wide expression profiles**. *Proc Natl Acad Sci U S A* 2005, **102**(43):15545-15550.
2. Huang da W, Sherman BT, Lempicki RA: **Bioinformatics enrichment tools: paths toward the comprehensive functional analysis of large gene lists**. *Nucleic Acids Res* 2009, **37**(1):1-13.
3. Huang da W, Sherman BT, Lempicki RA: **Systematic and integrative analysis of large gene lists using DAVID bioinformatics resources**. *Nat Protoc* 2009, **4**(1):44-57.
4. Berriz GF, King OD, Bryant B, Sander C, Roth FP: **Characterizing gene sets with FuncAssociate**. *Bioinformatics* 2003, **19**(18):2502-2504.
5. Chen J, Bardes EE, Aronow BJ, Jegga AG: **ToppGene Suite for gene list enrichment analysis and candidate gene prioritization**. *Nucleic Acids Res* 2009, **37**(Web Server issue):W305-311.
6. Kuleshov MV, Jones MR, Rouillard AD, Fernandez NF, Duan Q, Wang Z, Koplev S, Jenkins SL, Jagodnik KM, Lachmann A *et al*: **Enrichr: a comprehensive gene set enrichment analysis web server 2016 update**. *Nucleic Acids Res* 2016, **44**(W1):W90-97.
7. Cancer Genome Atlas N: **Comprehensive molecular portraits of human breast tumours**. *Nature* 2012, **490**(7418):61-70.
8. Chang YH, Chen CM, Chen HY, Yang PC: **Pathway-based gene signatures predicting clinical outcome of lung adenocarcinoma**. *Sci Rep* 2015, **5**:10979.
9. Tang H, Xiao G, Behrens C, Schiller J, Allen J, Chow CW, Suraokar M, Corvalan A, Mao J, White MA *et al*: **A 12-gene set predicts survival benefits from adjuvant chemotherapy in non-small cell lung cancer patients**. *Clin Cancer Res* 2013, **19**(6):1577-1586.
10. Lau SK, Boutros PC, Pintilie M, Blackhall FH, Zhu CQ, Strumpf D, Johnston MR, Darling G, Keshavjee S, Waddell TK *et al*: **Three-gene prognostic classifier for early-stage non small-cell lung cancer**. *J Clin Oncol* 2007, **25**(35):5562-5569.
11. Zeng H, Edwards MD, Liu G, Gifford DK: **Convolutional neural network architectures for predicting DNA-protein binding**. *Bioinformatics* 2016, **32**(12):i121-i127.
12. Arvaniti E, Claassen M: **Sensitive detection of rare disease-associated cell subsets via representation learning**. *Nat Commun* 2017, **8**:14825.
13. Lin C, Jain S, Kim H, Bar-Joseph Z: **Using neural networks for reducing the dimensions of single-cell RNA-Seq data**. *Nucleic Acids Res* 2017, **45**(17):e156.
14. Tan J, Hammond JH, Hogan DA, Greene CS: **ADAGE-Based Integration of Publicly Available Pseudomonas aeruginosa Gene Expression Data with Denoising Autoencoders Illuminates Microbe-Host Interactions**. *mSystems* 2016, **1**(1).
15. Chen Y, Li Y, Narayan R, Subramanian A, Xie X: **Gene expression inference with deep learning**. *Bioinformatics* 2016, **32**(12):1832-1839.

16. Newton Y, Novak AM, Swatloski T, McColl DC, Chopra S, Graim K, Weinstein AS, Baertsch R, Salama SR, Ellrott K et al: **TumorMap: Exploring the Molecular Similarities of Cancer Samples in an Interactive Portal**. *Cancer Res* 2017, **77**(21):e111-e114.
17. Liberzon A, Subramanian A, Pinchback R, Thorvaldsdottir H, Tamayo P, Mesirov JP: **Molecular signatures database (MSigDB) 3.0**. *Bioinformatics* 2011, **27**(12):1739-1740.
18. Hsiao TH, Chiu YC, Hsu PY, Lu TP, Lai LC, Tsai MH, Huang TH, Chuang EY, Chen Y: **Differential network analysis reveals the genome-wide landscape of estrogen receptor modulation in hormonal cancers**. *Sci Rep* 2016, **6**:23035.
19. He K, Zhang X, Ren S, Sun J: **Delving Deep into Rectifiers: Surpassing Human-Level Performance on ImageNet Classification**. In: *ArXiv e-prints.* vol. 1502; 2015.
20. van der Maaten LJP: **Accelerating t-SNE using Tree-Based Algorithms**. *Journal of Machine Learning Research* 2014, **15**:3221-3245.
21. van der Maaten LJP, Hinton GE: **Visualizing High-Dimensional Data Using t-SNE**. *Journal of Machine Learning Research* 2008, **9**:2579-2605.
22. Campello RJGB, Moulavi D, Sander J: **Density-Based Clustering Based on Hierarchical Density Estimates**. In*: 2013; Berlin, Heidelberg*: Springer Berlin Heidelberg; 2013: 160-172.
23. McInnes L, Healy J, Astels S: **hdbscan: Hierarchical density based clustering**. *The Journal of Open Source Software* 2017, **2**.
24. Farmer P, Bonnefoi H, Becette V, Tubiana-Hulin M, Fumoleau P, Larsimont D, Macgrogan G, Bergh J, Cameron D, Goldstein D et al: **Identification of molecular apocrine breast tumours by microarray analysis**. *Oncogene* 2005, **24**(29):4660-4671.
25. Stein RA, Chang CY, Kazmin DA, Way J, Schroeder T, Wergin M, Dewhirst MW, McDonnell DP: **Estrogen-related receptor alpha is critical for the growth of estrogen receptor-negative breast cancer**. *Cancer Res* 2008, **68**(21):8805-8812.
26. Doane AS, Danso M, Lal P, Donaton M, Zhang L, Hudis C, Gerald WL: **An estrogen receptor-negative breast cancer subset characterized by a hormonally regulated transcriptional program and response to androgen**. *Oncogene* 2006, **25**(28):3994-4008.
27. Pedersen K, Angelini PD, Laos S, Bach-Faig A, Cunningham MP, Ferrer-Ramon C, Luque-Garcia A, Garcia-Castillo J, Parra-Palau JL, Scaltriti M et al: **A naturally occurring HER2 carboxy-terminal fragment promotes mammary tumor growth and metastasis**. *Mol Cell Biol* 2009, **29**(12):3319-3331.
28. Thorner AR, Parker JS, Hoadley KA, Perou CM: **Potential tumor suppressor role for the c-Myb oncogene in luminal breast cancer**. *PLoS One* 2010, **5**(10):e13073.
29. Goldman M, Craft B, Swatloski T, Cline M, Morozova O, Diekhans M, Haussler D, Zhu J: **The UCSC Cancer Genomics Browser: update 2015**. *Nucleic Acids Res* 2015, **43**(Database issue):D812-817.

30. Zhu J, Sanborn JZ, Benz S, Szeto C, Hsu F, Kuhn RM, Karolchik D, Archie J, Lenburg ME, Esserman LJ et al: **The UCSC Cancer Genomics Browser**. *Nat Methods* 2009, **6**(4):239-240.
31. Pumperla M: **Keras + Hyperopt: A very simple wrapper for convenient hyperparameter optimization**. In.; 2016.
32. Landi MT, Dracheva T, Rotunno M, Figueroa JD, Liu H, Dasgupta A, Mann FE, Fukuoka J, Hames M, Bergen AW et al: **Gene expression signature of cigarette smoking and its role in lung adenocarcinoma development and survival**. *PLoS One* 2008, **3**(2):e1651.
33. Xiao J, Liu A, Lu X, Chen X, Li W, He S, He B, Chen Q: **Prognostic significance of TCF21 mRNA expression in patients with lung adenocarcinoma**. *Sci Rep* 2017, **7**(1):2027.
34. Zhang Y, Foreman O, Wigle DA, Kosari F, Vasmatzis G, Salisbury JL, van Deursen J, Galardy PJ: **USP44 regulates centrosome positioning to prevent aneuploidy and suppress tumorigenesis**. *J Clin Invest* 2012, **122**(12):4362-4374.
35. Qi XW, Zhang F, Wu H, Liu JL, Zong BG, Xu C, Jiang J: **Wilms' tumor 1 (WT1) expression and prognosis in solid cancer patients: a systematic review and meta-analysis**. *Sci Rep* 2015, **5**:8924.
36. Hiroumi H, Dosaka-Akita H, Yoshida K, Shindoh M, Ohbuchi T, Fujinaga K, Nishimura M: **Expression of E1AF/PEA3, an Ets-related transcription factor in human non-small-cell lung cancers: its relevance in cell motility and invasion**. *Int J Cancer* 2001, **93**(6):786-791.
37. Yamaguchi E, Nakayama T, Nanashima A, Matsumoto K, Yasutake T, Sekine I, Nagayasu T: **Ets-1 proto-oncogene as a potential predictor for poor prognosis of lung adenocarcinoma**. *Tohoku J Exp Med* 2007, **213**(1):41-50.
38. Baker AF, Hanke NT, Sands BJ, Carbajal L, Anderl JL, Garland LL: **Carfilzomib demonstrates broad anti-tumor activity in pre-clinical non-small cell and small cell lung cancer models**. *J Exp Clin Cancer Res* 2014, **33**:111.
39. Massuti B, Cobo M, Camps C, Domine M, Provencio M, Alberola V, Vinolas N, Rosell R, Taron M, Gutierrez-Calderon V et al: **Trabectedin in patients with advanced non-small-cell lung cancer (NSCLC) with XPG and/or ERCC1 overexpression and BRCA1 underexpression and pretreated with platinum**. *Lung Cancer* 2012, **76**(3):354-361.
40. Iwanaga K, Yang Y, Raso MG, Ma L, Hanna AE, Thilaganathan N, Moghaddam S, Evans CM, Li H, Cai WW et al: **Pten inactivation accelerates oncogenic K-ras-initiated tumorigenesis in a mouse model of lung cancer**. *Cancer Res* 2008, **68**(4):1119-1127.
41. Ooi AT, Mah V, Nickerson DW, Gilbert JL, Ha VL, Hegab AE, Horvath S, Alavi M, Maresh EL, Chia D et al: **Presence of a putative tumor-initiating progenitor cell population predicts poor prognosis in smokers with non-small cell lung cancer**. *Cancer Res* 2010, **70**(16):6639-6648.
42. Zhang W, Sun J, Luo J: **High Expression of Rab-like 3 (Rabl3) is Associated with Poor Survival of Patients with Non-Small Cell Lung Cancer via Repression of MAPK8/9/10-Mediated Autophagy**. *Med Sci Monit* 2016, **22**:1582-1588.

43. Han X, Wang R, Zhou Y, Fei L, Sun H, Lai S, Saadatpour A, Zhou Z, Chen H, Ye F *et al*: **Mapping the Mouse Cell Atlas by Microwell-Seq**. *Cell* 2018, **172**(5):1091-1107 e1017.
44. Rozenblatt-Rosen O, Stubbington MJT, Regev A, Teichmann SA: **The Human Cell Atlas: from vision to reality**. *Nature* 2017, **550**(7677):451-453.
45. Wang B, Zhu J, Pierson E, Ramazzotti D, Batzoglou S: **Visualization and analysis of single-cell RNA-seq data by kernel-based similarity learning**. *Nat Methods* 2017, **14**(4):414-416.
46. Chen HI, Jin Y, Huang Y, Chen Y: **Detection of high variability in gene expression from single-cell RNA-seq profiling**. *BMC Genomics* 2016, **17 Suppl 7**:508.

## Figures Legends

**Figure 1.** The architecture of gene superset autoencoder (GSAE). In the gene set layer, one color node represents a gene set, and edges in the same color show connect associate genes to a gene set.

**Figure 2.** The t-SNE results of TCGA 9,806 samples using (A) logTPM data with 15,975 genes (an initial PCA step was performed), and (B) 200 superset outputs.

**Figure 3.** Subtype analysis in BRCA data set. (A) The t-SNE results of BRCA data, where HDBSCAN classified the samples into two groups. The noisy samples were labeled in black and omitted from further analysis. (B) The density plots of the most significant up-superset and three selected top gene sets. The blue/yellow arrow corresponds to positive/negative weight in the model between the gene set and superset. (C) The density plots of the most significant down-superset and three selected top gene sets. (D) The Venn diagram of the significant gene sets in the top 3 up-supersets.

**Figure 4.** The Kaplan-Meier Curves of (A) $1^{st}$ ranked superset and selected three top 20 gene sets associated with the superset (B) $4^{th}$ ranked superset and selected three top 20 gene sets associated with the superset. The blue/yellow arrow corresponds to positive/negative weight in the model between the gene set and superset.

# Tables

**Table 1.** Top 15 gene sets in up-superset #1 in BRCA subtype analysis.

| Gene Set Terms | *PScore*[A] | *gsScore* | weight[B] |
|---|---|---|---|
| CUI_TCF21_TARGETS_2_UP | 81.999 | 0.980 | 0.198 |
| PEDERSEN_METASTASIS_BY_ERBB2_ISOFORM_7 | 92.578 | 0.927 | -0.154 |
| GRADE_COLON_AND_RECTAL_CANCER_UP | 68.314 | 0.427 | 0.138 |
| DOANE_BREAST_CANCER_ESR1_DN | 87.537 | 0.374 | 0.066 |
| VANTVEER_BREAST_CANCER_ESR1_DN | 80.186 | 0.366 | 0.083 |
| HATADA_METHYLATED_IN_LUNG_CANCER_UP | 76.111 | 0.333 | -0.103 |
| FARMER_BREAST_CANCER_BASAL_VS_LULMINAL | 90.606 | 0.321 | 0.079 |
| RICKMAN_TUMOR_DIFFERENTIATED_WELL_VS_MODERATELY_UP | 80.636 | 0.247 | -0.113 |
| BOQUEST_STEM_CELL_DN | 10.658 | 0.245 | -0.130 |
| BONOME_OVARIAN_CANCER_SURVIVAL_OPTIMAL_DEBULKING | 54.414 | 0.228 | -0.100 |
| MOREAUX_MULTIPLE_MYELOMA_BY_TACI_UP | 88.073 | 0.206 | -0.059 |
| YANG_BREAST_CANCER_ESR1_DN | 71.886 | 0.194 | 0.088 |
| DOUGLAS_BMI1_TARGETS_DN | 18.273 | 0.177 | -0.127 |
| STEIN_ESRRA_TARGETS_RESPONSIVE_TO_ESTROGEN_UP | 84.141 | 0.174 | -0.076 |
| BERNARD_PPAPDC1B_TARGETS_DN | 88.410 | 0.161 | -0.073 |

[A]The *PScore* of gene set Mann Whitney *U* test with location shift = 0.5.
[B]The weight in the model corresponding to the connection of a gene set to the corresponding superset.

**Table 2.** Top 15 gene sets in down-superset #1 in BRCA subtype analysis.

| Gene Set Terms | PScore[A] | gsScore | weight[B] |
|---|---:|---:|---:|
| PEDERSEN_METASTASIS_BY_ERBB2_ISOFORM_7 | 92.578 | 0.997 | 0.166 |
| VANTVEER_BREAST_CANCER_ESR1_DN | 80.186 | 0.811 | -0.185 |
| LEI_MYB_TARGETS | 55.903 | 0.800 | 0.201 |
| DOANE_BREAST_CANCER_ESR1_DN | 87.537 | 0.644 | -0.114 |
| CUI_TCF21_TARGETS_2_UP | 81.999 | 0.511 | -0.103 |
| ACEVEDO_NORMAL_TISSUE_ADJACENT_TO_LIVER_TUMOR_DN | 46.415 | 0.340 | 0.127 |
| DELACROIX_RARG_BOUND_MEF | 55.442 | 0.336 | 0.141 |
| SENGUPTA_NASOPHARYNGEAL_CARCINOMA_UP | 37.512 | 0.230 | -0.074 |
| FRASOR_RESPONSE_TO_ESTRADIOL_DN | 77.02 | 0.218 | 0.108 |
| HATADA_METHYLATED_IN_LUNG_CANCER_UP | 76.111 | 0.215 | 0.066 |
| SMID_BREAST_CANCER_LUMINAL_A_UP | 51.385 | 0.211 | 0.077 |
| FOSTER_KDM1A_TARGETS_UP | 62.021 | 0.188 | 0.091 |
| SHEDDEN_LUNG_CANCER_GOOD_SURVIVAL_A12 | 6.764 | 0.178 | -0.096 |
| SWEET_LUNG_CANCER_KRAS_UP | 6.548 | 0.178 | 0.138 |
| KOYAMA_SEMA3B_TARGETS_DN | 24.534 | 0.176 | 0.094 |

[A]The *PScore* of gene set Mann Whitney *U* test.
[B]The weight in the model corresponding to the connection of a gene set to the superset.

**Table 3.** The size of encoder layers and the 10-fold cross validation accuracy of each neural network classifier.

| NN classifier [A] | Input Type | Encoder Layer 1 [D] | Encoder Layer 2 | Encoder Layer 3 | Encoder Layer 4 | Accuracy of 10-fold cross validation |
|---|---|---|---|---|---|---|
| Superset | Genes [B] | 2,337 | 200 | | | 88.79% |
| Gene set | Genes | 2,337 | | | | 87.69% |
| 2-layer fc | Genes | 2,337 | 200 | | | 47.86% |
| 2-layer fc | Genes | 2,000 | 500 | | | 37.98% |
| 4-layer fc | Genes | 2,000 | 200 | 100 | 50 | 46.06% |
| 2-layer fc | PC [C] | 400 | 100 | | | 87.57% |
| 4-layer fc | PC | 200 | 200 | 100 | 25 | 87.57% |

[A] 2-, 4-layer fc: 2- or 4- layer fully connected AE.
[B] Genes input is the 15,183 genes of TCGA BRCA RNA-seq data.
[C] PC Input is the top 500 principal components of PCA analysis.
[D] The encoder layer 1 of superset and gene set classifier is the gene set layer (not a fully connected layer).

**Table 4.** Top 15 gene sets in the highest ranked superset in LUAD survival analysis.

| Gene Set Terms | $P$-value[A] | gsScore | weight[B] |
|---|---|---|---|
| CUI_TCF21_TARGETS_2_UP | $1.30 \times 10^{-4}$ | 0.446 | -0.211 |
| RICKMAN_TUMOR_DIFFERENTIATED_WELL_VS_POORLY_DN | $2.02 \times 10^{-4}$ | 0.254 | -0.201 |
| GROSS_HYPOXIA_VIA_ELK3_UP | $5.21 \times 10^{-4}$ | 0.245 | 0.115 |
| KAAB_HEART_ATRIUM_VS_VENTRICLE_DN | $5.10 \times 10^{-4}$ | 0.194 | 0.145 |
| MARTINEZ_RESPONSE_TO_TRABECTEDIN_DN | 0.0015 | 0.183 | 0.107 |
| MITSIADES_RESPONSE_TO_APLIDIN_UP | 0.2863 | 0.159 | -0.171 |
| KIM_WT1_TARGETS_DN | 0.0064 | 0.146 | 0.081 |
| ENK_UV_RESPONSE_EPIDERMIS_UP | $1 \times 10^{-5}$ | 0.143 | 0.077 |
| SENESE_HDAC1_TARGETS_DN | 0.8285 | 0.138 | -0.162 |
| SENGUPTA_NASOPHARYNGEAL_CARCINOMA_WITH_LMP1_UP | 0.1411 | 0.129 | -0.130 |
| YANG_BCL3_TARGETS_UP | 0.0299 | 0.126 | 0.163 |
| GINESTIER_BREAST_CANCER_ZNF217_AMPLIFIED_DN | 0.9507 | 0.124 | 0.132 |
| CONCANNON_APOPTOSIS_BY_EPOXOMICIN_DN | 0.0264 | 0.112 | 0.147 |
| SPIELMAN_LYMPHOBLAST_EUROPEAN_VS_ASIAN_UP | 0.0048 | 0.110 | -0.051 |
| RHEIN_ALL_GLUCOCORTICOID_THERAPY_DN | 0.0154 | 0.107 | 0.051 |

[A]The $P$-value of gene set log-rank.
[B]The weight in the model corresponding to the connection of a gene set to the superset.

**Table 5.** Top 15 gene sets in 4th ranked superset in LUAD survival analysis.

| Gene Set Terms | P-value[A] | gsScore | weight[B] |
|---|---|---|---|
| SWEET_LUNG_CANCER_KRAS_DN | 0.7304 | 0.780 | -0.185 |
| ZHANG_BREAST_CANCER_PROGENITORS_UP | 0.0248 | 0.256 | 0.096 |
| ROZANOV_MMP14_TARGETS_UP | 0.1038 | 0.161 | 0.103 |
| MONNIER_POSTRADIATION_TUMOR_ESCAPE_DN | 0.0058 | 0.157 | -0.117 |
| ACEVEDO_FGFR1_TARGETS_IN_PROSTATE_CANCER_MODEL_DN | 0.0988 | 0.154 | 0.114 |
| YOSHIMURA_MAPK8_TARGETS_DN | 0.0195 | 0.150 | -0.126 |
| DELYS_THYROID_CANCER_DN | 0.0065 | 0.125 | -0.079 |
| SWEET_LUNG_CANCER_KRAS_UP | 0.2762 | 0.122 | 0.141 |
| OSWALD_HEMATOPOIETIC_STEM_CELL_IN_COLLAGEN_GEL_DN | 0.0132 | 0.101 | 0.120 |
| GROSS_HYPOXIA_VIA_ELK3_UP | $5.21 \times 10^{-4}$ | 0.100 | 0.058 |
| WATTEL_AUTONOMOUS_THYROID_ADENOMA_UP | 0.1555 | 0.096 | -0.089 |
| VERHAAK_GLIOBLASTOMA_MESENCHYMAL | 0.7314 | 0.095 | 0.113 |
| PHONG_TNF_RESPONSE_NOT_VIA_P38 | 0.0972 | 0.093 | -0.121 |
| RUTELLA_RESPONSE_TO_HGF_UP | 0.7217 | 0.091 | -0.055 |
| IWANAGA_CARCINOGENESIS_BY_KRAS_PTEN_UP | 0.0249 | 0.090 | 0.088 |

[A]The *P*-value of gene set log-rank.
[B]The weight in the model corresponding to the connection of a gene set to the superset.

**Table 6.** The statistical information of GSAE outputs between the training and test TCGA data sets of four cancer types.

| TCGA data set | Superset Jaccard Index[A] | Gene set Jaccard Index[B] | Two proportion z-test | | |
|---|---|---|---|---|---|
| | | | Superset Proportion[C] | Gene set Proportion[D] | $P$-value[E] |
| BRCA | 34.38% | 12.35% | 11 / 24 | 31 / 197 | 0.0002 |
| LUAD | 18.18% | 11.27% | 6 / 12 | 32 / 145 | 0.0150 |
| SKCM | 17.86% | 6.88% | 5 / 19 | 17 / 139 | 0.0485 |
| LGG | 48.33% | 47.54% | 29 / 45 | 299 / 481 | 0.3821 |

Supersets/gene sets with log-rank $P$-value < 0.05 were selected as prognostic significant sets.
[A]Jaccard index of significant supersets between training and test data.
[B]Jaccard index of significant gene sets between training and test data.
[C]Superset proportion: (# of overlapped significant supersets between training and test data) over (# of significant supersets in training data).
[D]Gene set proportion: (# of overlapped significant gene sets between training and test data) over (# of significant gene sets in training data).
[E]The $P$-value of z-test comparing superset and gene set proportions.

## Additional files

**Additional file 1 – Fig_S1.pdf**

The architectures of four neural network classifiers. (A) superset classifier, (B) gene set classifier, (C) 2-layer fully connected encoder network classifier, and (D) 4-layer fully connected encoder classifier.

**Additional file 2 – Table_S1.xlsx**

Top high impact gene sets of the four up-supersets and three down-supersets in the BRCA tumor subtype analysis. The highly overlapped collections of a gene set were determined by MSigDB (in the compute overlaps section).

**Additional file 3 – Table_S2.xlsx**

Top 20 gene sets of the six prognostic significant supersets in LUAD survival analysis.

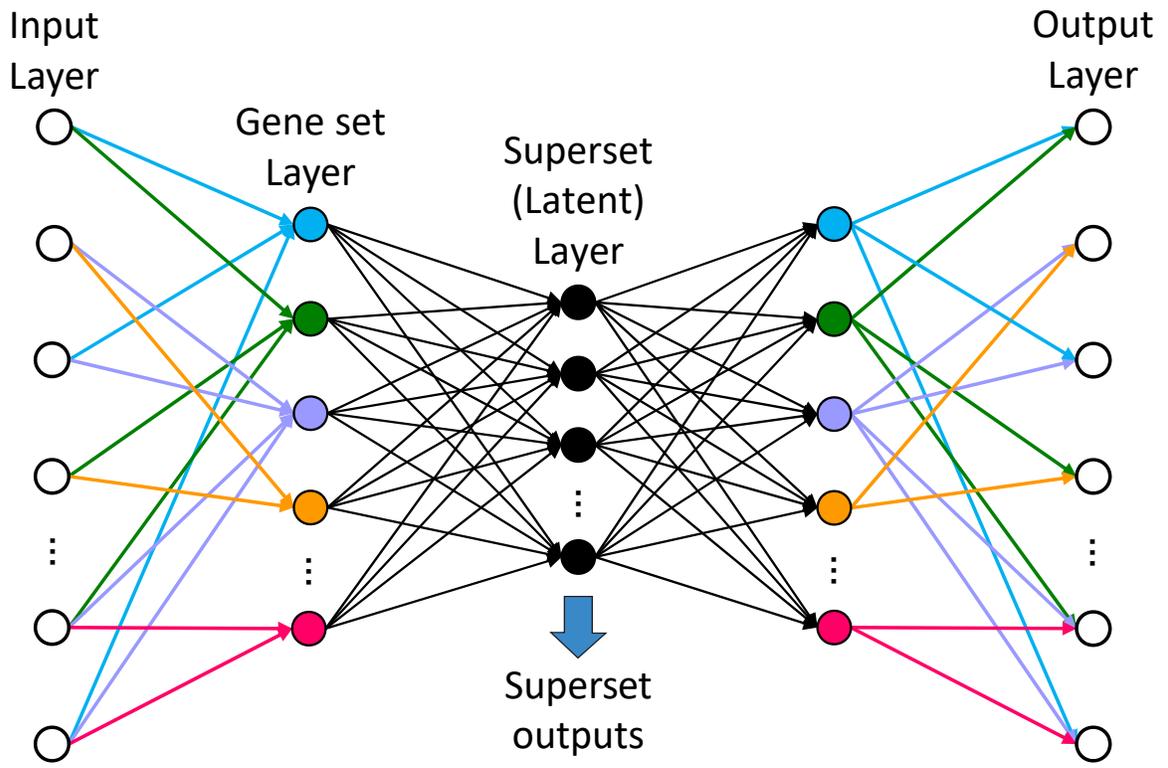

Figure 1

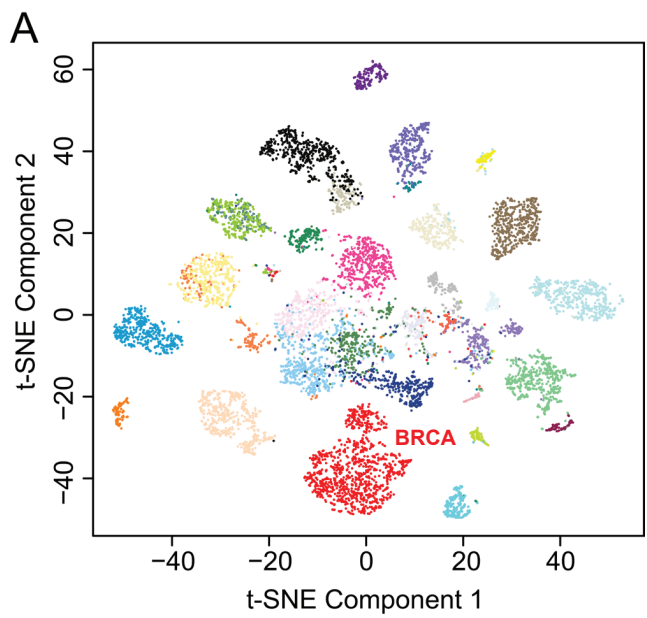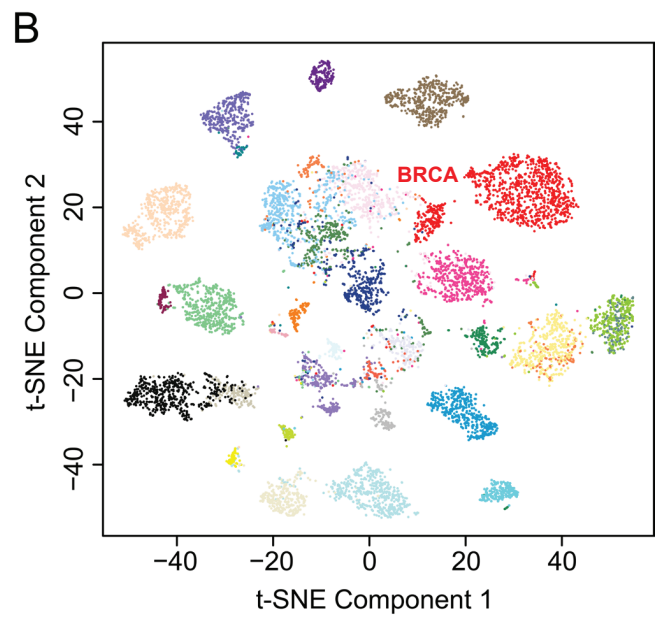

Figure 2

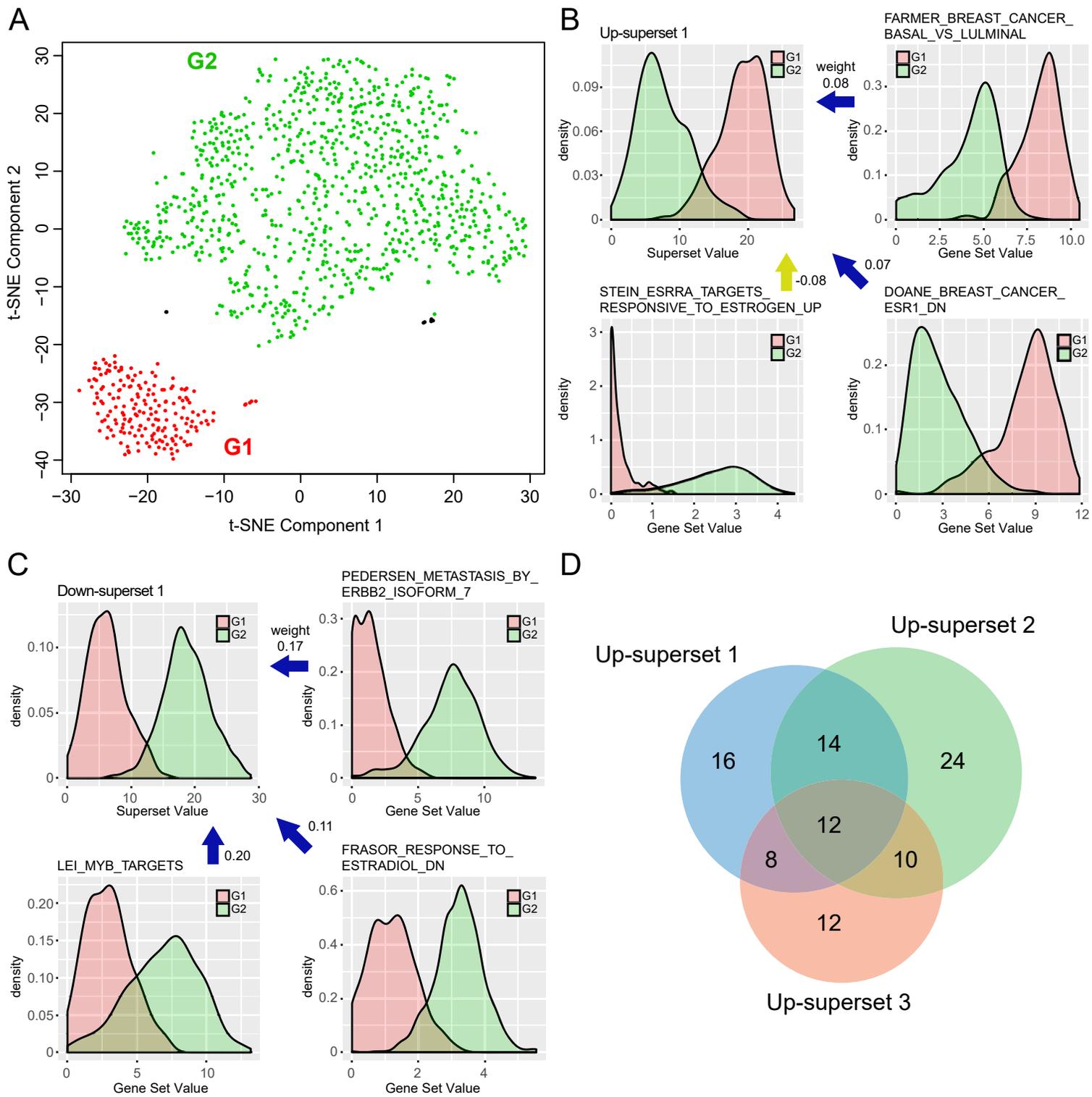

Figure 3

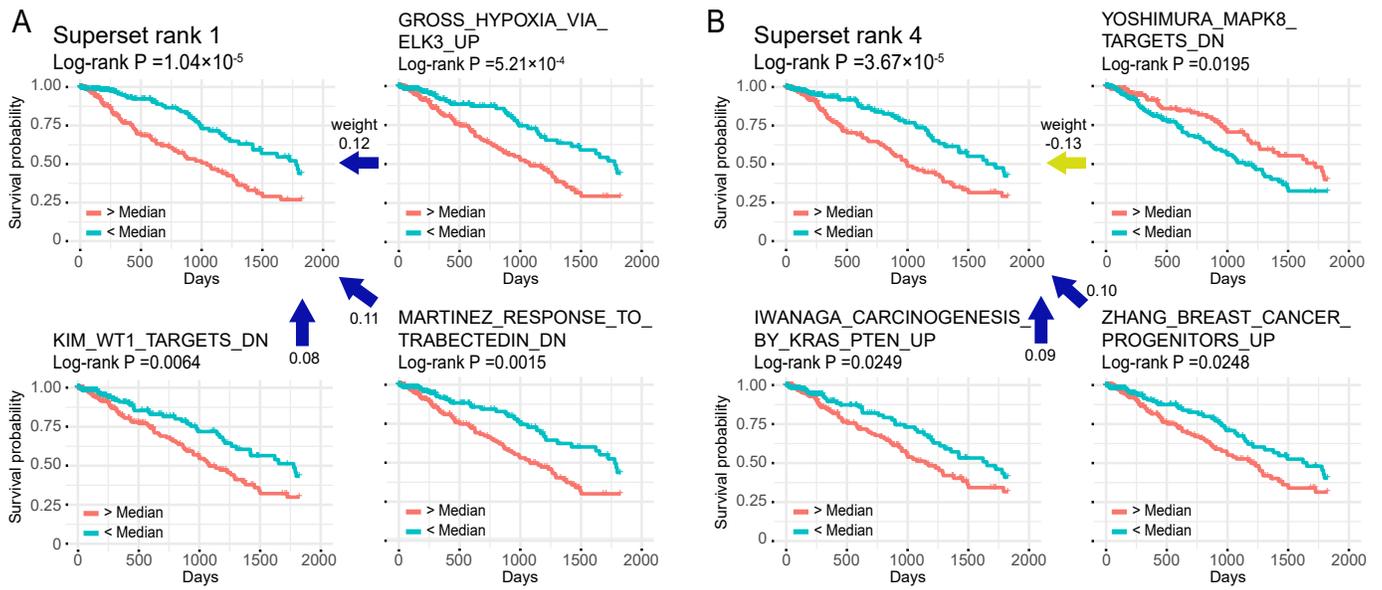

Figure 4